\documentclass{article}

\usepackage{microtype}
\usepackage{graphicx}
\usepackage{subfigure}
\usepackage{booktabs} %

\usepackage[hyphenbreaks]{breakurl}
\PassOptionsToPackage{hyphens}{url}
\usepackage{hyperref}

\usepackage[accepted]{icml2023}

\usepackage{amsmath}
\usepackage{amssymb}
\usepackage{mathtools}
\usepackage{amsthm}

\usepackage[capitalize,noabbrev]{cleveref}

\theoremstyle{plain}

\theoremstyle{definition}

\theoremstyle{remark}

\usepackage[textsize=tiny]{todonotes}

\usepackage{paralist}
\usepackage[marginal,hang,stable]{footmisc}

\newcommand{\descr}[1]{\noindent\textbf{#1}}
\icmltitlerunning{When Synthetic Data Met Regulation}

\begin{document}

\twocolumn[
\icmltitle{When Synthetic Data Met Regulation}

\begin{icmlauthorlist}
  \icmlauthor{Georgi Ganev}{hazy,ucl}
\end{icmlauthorlist}

\icmlaffiliation{hazy}{Hazy, London, UK}
\icmlaffiliation{ucl}{University College London, London, UK}

\icmlcorrespondingauthor{Georgi Ganev}{georgi.ganev.16@ucl.ac.uk}

\icmlkeywords{Synthetic Data, Regulation, Differential Privacy, Generative AI, Machine Learning, GenLaw, ICML}

\vskip 0.3in
]

\printAffiliationsAndNotice{}  %

\section{Motivation}
\vspace{-0.1cm}

Generative AI has made significant progress recently, with applications spanning text, code, image, video, speech, and structured data~\cite{sequoia2022generative}.
Investor interest has also grown -- start-ups received \$2.2B in 2022~\cite{techcrunch2023vcs} and Microsoft reportedly invested \$10B in OpenAI's ChatGPT~\cite{bloomberg2023microsoft}, which has reached 100M monthly users~\cite{reuters2023chatgpt}.
However, concerns about privacy, robustness, copyright, and compliance have increased as well.
Active legal cases against Generative AI companies and products~\cite{techcrunch2023the} have led some organizations and countries, such as Italy, to (temporarily) restrict ChatGPT usage~\cite{cnn2023dont, politico2023italian}.

\descr{Synthetic Data.}
In this paper, we focus on synthetic data, a subfield of Generative AI that utilizes generative machine learning models such as GANs~\cite{goodfellow2014generative}, Diffusion Models~\cite{sohl2015deep}, and Transformers~\cite{vaswani2017attention}, albeit typically at a smaller scale.
We opt for tabular data comprising sensitive information as training data as it is still the most extensively used data type in large enterprises.
Furthermore, synthetic data is comparatively more established and has recently been examined by reputable organizations~\cite{rs2023privacy, un2023guide} and regulators~\cite{ico2022privacy, fca2023synthetic}, alas without any definitive compliance directives.

\descr{Main Question.}
This prompts the question: {\em ``Can we make synthetic data regulatory compliant?''}
Namely, we explore the legality of privacy-preserving synthetic data created by generative models trained on structured personal data.

\vspace{-0.15cm}
\section{Regulatory Definitions}
\vspace{-0.1cm}

\descr{Personal Data.}
\citet{official2016article} define personal data as ``any information relating to an identified or identifiable living individual'' and the latter as someone who can be identified (directly or indirectly) by reference to factors such as name, id number, or physical, genetic, social identity, etc.
On the other hand, information that is effectively anonymized is not personal data and data protection law does not apply to it~\cite{official2016recital}.
But in practice the actual identifiability of individuals can be highly context-specific as different types of information carry different levels of identifiability risks depending on the circumstances.
Clearly, creating synthetic data based on sensitive personal data requires processing it.
However, whether the resultant synthetic data constitutes personal or anonymous information is a question to be determined based on an assessment of the identifiability risk.
This raises the question, what constitutes a sufficient level of anonymization.

\descr{Sufficient Anonymization.}
\citet{ico2021how} states that ``effective anonymization reduces identifiability risk to a sufficiently remote level.''
When assessing whether someone is identifiable, objective factors to be considered include the cost and time required to identify, the available technologies, and their developments over time.
However, not every hypothetical/theoretical chance of identifiability needs to be taken into account.
The focus should be on what is reasonably likely to be used relative to the circumstances, not in absolute.
This is consistent with \citet{eu2014opinion}'s approach, that also notes that data controllers should regularly reassess the attending risks.
In terms of technical analysis, \citet{eu2014opinion, ico2021how} assert that the following three key risks need to be reduced for sufficient anonymization:
\begin{compactenum}
  \item ({\em singling out}) any individual being isolated;
  \item ({\em linkability}) any records/datasets (publicly available or not) being combined with synthetic data and thereby enabling the identification of an individual;
  \item ({\em inferences}) an attribute being deduced with significant probability from the values of other attributes.\footnote{This is in direct contradiction with good quality synthetic data and has led to leading privacy researchers abandoning statistical inference as privacy violation~\cite{mcsherry2016statistical, bun2021statistical}.}
\end{compactenum}

\citet{ico2021how} explains that the three risks should be looked through the {\em motivated intruder} test -- a competent intruder having access to appropriate resources being able to achieve identification if they were motivated to attempt it.

\vspace{-0.15cm}
\section{Synthetic Data as Anonymous Data}
\vspace{-0.1cm}

In this section, we show that producing synthetic data by combining two techniques---generative models and Differential Privacy (DP)---reduces all identifiability risks to sufficiently remote level and, therefore, the resulting data can be considered anonymous per~\cite{eu2014opinion, ico2021how}.
Overall, we rely on generative models to create high utility synthetic data and DP to provably guarantee privacy.

\descr{Generative Models}
break the 1-to-1 mapping and to an extent reduce singling out and linkability but could be susceptible to various privacy attacks (see below).

The process of training a generative model to learn the probability distribution of the input sensitive data, discarding it, and sampling from the fitted parameters to create new (synthetic) data, naturally lowers some privacy concerns.
For instance, it breaks the 1-to-1 mapping from a single real record to a single synthetic one which makes singling out difficult.
Since the models are probabilistic in nature, they capture the inherent data uncertainty and variability, which reduces linkability.
Furthermore, launching adversarial privacy attacks versus generative models is more challenging compared to discriminative ones~\cite{de2021critical}.

However, some generative models could occasionally memorize records and reproduce them (exactly or approximately) in the synthetic data~\cite{carlini2019secret, van2021memorization}.
In turn, a strategic adversary with side knowledge (e.g., the training algorithm, representable data, etc.) could infer the presence of these records~\cite{hayes2019logan, chen2020gan, stadler2022synthetic}, thus violating the linkability test and rendering the synthetic data pseudonymous at best or personal at worst~\cite{lopez2022on}.
Even more powerful privacy attack is reconstruction~\cite{carlini2021extracting, carlini2023extracting}, in which the adversary manages to recover whole training records and, therefore, leaks all of their private attributes.

\descr{DP}
mechanisms formally protect against singling out, linkability, and other re-identifiability concerns even if faced with a resourceful and strategic adversary (see below).

DP~\cite{dwork2006calibrating, dwork2014algorithmic} is a mathematical definition of privacy which formally bounds the probability of distinguishing whether any given individual’s data was included in the input data.
The level of indistinguishability is controlled and quantified by a parameter, $\epsilon$, or the privacy budget.
In the context of Generative AI, DP is usually satisfied by training the models with noisy/random mechanisms and frameworks such as DP-SGD~\cite{abadi2016deep} and PATE~\cite{papernot2017semi, papernot2018scalable}.

Since DP makes the trained model indistinguishable, whether any individual’s data was included or not, it averts memorization and singling out.
The protection against GDPR’s singling out has been robustly formalized~\cite{cohen2020towards} (\citet{nissim2017bridging} also argue DP satisfies FERPA requirements).
Additionally, DP defends against potential harms, such as linkability, that could be caused by the publication of other sensitive information.
\citet{stadler2022synthetic} show this holds true even for outliers or potentially the most vulnerable individuals who have a higher chance of being memorized~\cite{feldman2020does}.
Furthermore, DP does not make any assumptions about the adversary and the provable mathematical guarantees apply in the worst-case scenario (e.g., the attacker has prior information, knowledge of the training algorithm, strong computing power, etc.) which means that DP protects against motivated adversaries.
The protections are not just theoretical, DP reduces all key risks empirically, too~\cite{giomi2022unified}.

Using DP-trained models makes privacy an attribute of the generating process rather than a given synthetic dataset.
Thanks to its resistance to post-processing property, DP allows reusing models (to generate data) without further privacy leakage.
This means that even in the unlikely scenario in which a synthetic record very similar to a real is generated (which could be dissatisfactory~\cite{ons2018privacy}), it does not constitute a privacy violation~\cite{jordon2022synthetic}.

\descr{Potential Limitations.}
While DP offers robust privacy protection, in certain scenarios it could be too conservative~\cite{nasr2021adversary}.
Furthermore, DP often leads to utility reduction, particularly impacting outliers and underrepresented subgroups~\cite{stadler2022synthetic, ganev2022robin} and causing inconsistencies~\cite{kulynych2023arbitrary}.
Selecting both the right privacy budget and DP mechanism is non-trivial and highly context-specific~\cite{hsu2014differential, ganev2023understanding}.
Lastly, implementing DP in practice and effectively conveying its properties can be challenging/complex~\cite{cummings2021need, houssiau2022on}.

\descr{Related Work.}
\citet{cummings2023challenges} discuss further DP benefits/challenges/open questions and \citet{jordon2022synthetic, de2023synthetic} focus on combining synthetic data with DP (also advised by~\cite{bellovin2019privacy}).
Specific (DP) generative models include GANs~\cite{xie2018differentially, jordon2018pate, xu2023synthetic}, Diffusion Models~\cite{kotelnikov2022tabddpm, ghalebikesabi2023differentially}, and Transformers~\cite{borisov2022language, solatorio2023realtabformer}.

\vspace{-0.15cm}
\section{Future Work}
\vspace{-0.1cm}

In this paper, we argue that synthetic data produced by DP generative models can be sufficiently anonymized and, therefore, anonymous data and regulatory compliant.
Our work aims to establish a foundation for broader Generative AI solutions.
Nevertheless, they face added obstacles, such as training on vast multi-modal datasets that may include proprietary/copyrighted data with commercial usage limitations.
Moreover, as datasets are often distributed over the internet, it becomes increasingly difficult for individuals to assert their right to consent or be forgotten.
Factors like data accessibility (e.g., decentralized/scraped data), governance, robustness, transparency, explainability, and fairness must also be considered~\cite{gal2023bridging, iapp2023generative}.

\vspace{-0.15cm}
\section*{Acknowledgements}
\vspace{-0.1cm}

We would like to thank Diana Sofronieva for helping with the structure, clarity, and rigorousness of the paper;
Ian Stevens, Samikah Ahmed, and Suzanne Jopling for providing legal expertise and consultation;
Emiliano De Cristofaro and Meenatchi Sundaram Muthu Selva Annamalai for providing technical feedback;
Adriano Basso, Andrew Keen, and Gareth Rees for reviewing earlier versions of the paper and offering overall support;
Orla Lynskey and James Jordon for the helpful discussions;
as well as the anonymous reviewers for their encouraging and helpful comments.

{\small

\begin{thebibliography}{59}
\providecommand{\natexlab}[1]{#1}
\providecommand{\url}[1]{\texttt{#1}}
\expandafter\ifx\csname urlstyle\endcsname\relax
  \providecommand{\doi}[1]{doi: #1}\else
  \providecommand{\doi}{doi: \begingroup \urlstyle{rm}\Url}\fi

\bibitem[{A29WP}(2014)]{eu2014opinion}
{A29WP}.
\newblock {Opinion on anonymisation techniques}.
\newblock
  \url{https://ec.europa.eu/justice/article-29/documentation/opinion-recommendation/files/2014/wp216_en.pdf},
  2014.

\bibitem[Abadi et~al.(2016)Abadi, Chu, Goodfellow, McMahan, Mironov, Talwar,
  and Zhang]{abadi2016deep}
Abadi, M., Chu, A., Goodfellow, I., McMahan, H.~B., Mironov, I., Talwar, K.,
  and Zhang, L.
\newblock {Deep learning with differential privacy}.
\newblock In \emph{ACM CCS}, 2016.

\bibitem[Bellovin et~al.(2019)Bellovin, Dutta, and
  Reitinger]{bellovin2019privacy}
Bellovin, S.~M., Dutta, P.~K., and Reitinger, N.
\newblock {Privacy and synthetic datasets}.
\newblock \emph{STLR}, 2019.

\bibitem[{Bloomberg}(2023)]{bloomberg2023microsoft}
{Bloomberg}.
\newblock {Microsoft Invests \$10 Billion in ChatGPT Maker OpenAI}.
\newblock
  \url{https://www.bloomberg.com/news/articles/2023-01-23/microsoft-makes-multibillion-dollar-investment-in-openai},
  2023.

\bibitem[Borisov et~al.(2022)Borisov, Se{\ss}ler, Leemann, Pawelczyk, and
  Kasneci]{borisov2022language}
Borisov, V., Se{\ss}ler, K., Leemann, T., Pawelczyk, M., and Kasneci, G.
\newblock {Language models are realistic tabular data generators}.
\newblock \emph{arXiv:2210.06280}, 2022.

\bibitem[Bun et~al.(2021)Bun, Desfontaines, Dwork, Naor, Nissim, Roth, Smith,
  Steinke, Ullman, and Vadhan]{bun2021statistical}
Bun, M., Desfontaines, D., Dwork, C., Naor, M., Nissim, K., Roth, A., Smith,
  A., Steinke, T., Ullman, J., and Vadhan, S.
\newblock {Statistical Inference is Not a Privacy Violation}.
\newblock
  \url{https://differentialprivacy.org/inference-is-not-a-privacy-violation/},
  2021.

\bibitem[Carlini et~al.(2019)Carlini, Liu, Erlingsson, Kos, and
  Song]{carlini2019secret}
Carlini, N., Liu, C., Erlingsson, {\'U}., Kos, J., and Song, D.
\newblock {The secret sharer: Evaluating and testing unintended memorization in
  neural networks}.
\newblock In \emph{USENIX Security}, 2019.

\bibitem[Carlini et~al.(2021)Carlini, Tramer, Wallace, Jagielski, Herbert-Voss,
  Lee, Roberts, Brown, Song, Erlingsson, Oprea, and
  Raffel]{carlini2021extracting}
Carlini, N., Tramer, F., Wallace, E., Jagielski, M., Herbert-Voss, A., Lee, K.,
  Roberts, A., Brown, T., Song, D., Erlingsson, U., Oprea, A., and Raffel, C.
\newblock {Extracting training data from large language models}.
\newblock In \emph{USENIX Security}, 2021.

\bibitem[Carlini et~al.(2023)Carlini, Hayes, Nasr, Jagielski, Sehwag,
  Tram{\`e}r, Balle, Ippolito, and Wallace]{carlini2023extracting}
Carlini, N., Hayes, J., Nasr, M., Jagielski, M., Sehwag, V., Tram{\`e}r, F.,
  Balle, B., Ippolito, D., and Wallace, E.
\newblock {Extracting training data from diffusion models}.
\newblock \emph{arXiv:2301.13188}, 2023.

\bibitem[Chen et~al.(2020)Chen, Yu, Zhang, and Fritz]{chen2020gan}
Chen, D., Yu, N., Zhang, Y., and Fritz, M.
\newblock {Gan-leaks: a taxonomy of membership inference attacks against
  generative models}.
\newblock In \emph{ACM CCS}, 2020.

\bibitem[{CNN}(2023)]{cnn2023dont}
{CNN}.
\newblock {Don’t tell anything to a chatbot you want to keep private}.
\newblock
  \url{https://edition.cnn.com/2023/04/06/tech/chatgpt-ai-privacy-concerns/index.html},
  2023.

\bibitem[Cohen \& Nissim(2020)Cohen and Nissim]{cohen2020towards}
Cohen, A. and Nissim, K.
\newblock {Towards formalizing the GDPR’s notion of singling out}.
\newblock \emph{PNAS}, 2020.

\bibitem[Cummings et~al.(2021)Cummings, Kaptchuk, and
  Redmiles]{cummings2021need}
Cummings, R., Kaptchuk, G., and Redmiles, E.~M.
\newblock {"I need a better description": an investigation into user
  expectations for differential privacy}.
\newblock In \emph{ACM CCS}, 2021.

\bibitem[Cummings et~al.(2023)Cummings, Desfontaines, Evans, Geambasu,
  Jagielski, Huang, Kairouz, Kamath, Oh, Ohrimenko, Papernot, Rogers, Shen,
  Song, Su, Terzis, Thakurta, Vassilvitskii, Wang, Xiong, Yekhanin, Yu, Zhan,
  and Zhang]{cummings2023challenges}
Cummings, R., Desfontaines, D., Evans, D., Geambasu, R., Jagielski, M., Huang,
  Y., Kairouz, P., Kamath, G., Oh, S., Ohrimenko, O., Papernot, N., Rogers, R.,
  Shen, M., Song, S., Su, W., Terzis, A., Thakurta, A., Vassilvitskii, S.,
  Wang, Y.-X., Xiong, L., Yekhanin, S., Yu, D., Zhan, H., and Zhang, W.
\newblock {Challenges towards the Next Frontier in Privacy}.
\newblock \emph{arXiv:2304.06929}, 2023.

\bibitem[De~Cristofaro(2021)]{de2021critical}
De~Cristofaro, E.
\newblock A critical overview of privacy in machine learning.
\newblock \emph{IEEE S\&P}, 2021.

\bibitem[De~Cristofaro(2023)]{de2023synthetic}
De~Cristofaro, E.
\newblock {What Is Synthetic Data? The Good, The Bad, and The Ugly}.
\newblock \emph{arXiv:2303.01230}, 2023.

\bibitem[Dwork \& Roth(2014)Dwork and Roth]{dwork2014algorithmic}
Dwork, C. and Roth, A.
\newblock {The algorithmic foundations of differential privacy}.
\newblock \emph{Foundations and Trends in Theoretical Computer Science}, 2014.

\bibitem[Dwork et~al.(2006)Dwork, McSherry, Nissim, and
  Smith]{dwork2006calibrating}
Dwork, C., McSherry, F., Nissim, K., and Smith, A.
\newblock {Calibrating noise to sensitivity in private data analysis}.
\newblock In \emph{TCC}, 2006.

\bibitem[{EP and Council}(2016{\natexlab{a}})]{official2016article}
{EP and Council}.
\newblock {Article 4 GDPR Definitions}.
\newblock \url{https://gdpr-info.eu/art-4-gdpr/}, 2016{\natexlab{a}}.

\bibitem[{EP and Council}(2016{\natexlab{b}})]{official2016recital}
{EP and Council}.
\newblock {Recital 26 EU GDPR}.
\newblock \url{https://www.privacy-regulation.eu/en/recital-26-GDPR.htm},
  2016{\natexlab{b}}.

\bibitem[{FCA UK}(2023)]{fca2023synthetic}
{FCA UK}.
\newblock {Synthetic data call for input feedback statement}.
\newblock \url{https://www.fca.org.uk/publication/feedback/fs23-1.pdf}, 2023.

\bibitem[Feldman(2020)]{feldman2020does}
Feldman, V.
\newblock {Does learning require memorization? a short tale about a long tail}.
\newblock In \emph{STOC}, 2020.

\bibitem[Gal \& Lynskey(2023)Gal and Lynskey]{gal2023bridging}
Gal, M. and Lynskey, O.
\newblock {Synthetic Data: Legal Implications of the Data-Generation
  Revolution}.
\newblock \emph{109 Iowa Law Review}, 2023.

\bibitem[Ganev et~al.(2022)Ganev, Oprisanu, and De~Cristofaro]{ganev2022robin}
Ganev, G., Oprisanu, B., and De~Cristofaro, E.
\newblock {Robin Hood and Matthew Effects: Differential privacy has disparate
  impact on synthetic data}.
\newblock In \emph{ICML}, 2022.

\bibitem[Ganev et~al.(2023)Ganev, Xu, and
  De~Cristofaro]{ganev2023understanding}
Ganev, G., Xu, K., and De~Cristofaro, E.
\newblock {Understanding how Differentially Private Generative Models Spend
  their Privacy Budget}.
\newblock \emph{arXiv:2305.10994}, 2023.

\bibitem[Ghalebikesabi et~al.(2023)Ghalebikesabi, Berrada, Gowal, Ktena,
  Stanforth, Hayes, De, Smith, Wiles, and
  Balle]{ghalebikesabi2023differentially}
Ghalebikesabi, S., Berrada, L., Gowal, S., Ktena, I., Stanforth, R., Hayes, J.,
  De, S., Smith, S.~L., Wiles, O., and Balle, B.
\newblock {Differentially Private Diffusion Models Generate Useful Synthetic
  Images}.
\newblock \emph{arXiv:2302.13861}, 2023.

\bibitem[Giomi et~al.(2022)Giomi, Boenisch, Wehmeyer, and
  Tasn{\'a}di]{giomi2022unified}
Giomi, M., Boenisch, F., Wehmeyer, C., and Tasn{\'a}di, B.
\newblock {A unified framework for quantifying privacy risk in synthetic data}.
\newblock In \emph{PETs}, 2022.

\bibitem[Goodfellow et~al.(2014)Goodfellow, Pouget-Abadie, Mirza, Xu,
  Warde-Farley, Ozair, Courville, and Bengio]{goodfellow2014generative}
Goodfellow, I., Pouget-Abadie, J., Mirza, M., Xu, B., Warde-Farley, D., Ozair,
  S., Courville, A., and Bengio, Y.
\newblock {Generative adversarial nets}.
\newblock \emph{NIPS}, 2014.

\bibitem[Hayes et~al.(2019)Hayes, Melis, Danezis, and
  De~Cristofaro]{hayes2019logan}
Hayes, J., Melis, L., Danezis, G., and De~Cristofaro, E.
\newblock {Logan: membership inference attacks against generative models}.
\newblock In \emph{PoPETs}, 2019.

\bibitem[Houssiau et~al.(2022)Houssiau, Rocher, and
  de~Montjoye]{houssiau2022on}
Houssiau, F., Rocher, L., and de~Montjoye, Y.-A.
\newblock {On the difficulty of achieving differential privacy in practice:
  user-level guarantees in aggregate location data}.
\newblock \emph{Nature Communications}, 2022.

\bibitem[Hsu et~al.(2014)Hsu, Gaboardi, Haeberlen, Khanna, Narayan, Pierce, and
  Roth]{hsu2014differential}
Hsu, J., Gaboardi, M., Haeberlen, A., Khanna, S., Narayan, A., Pierce, B.~C.,
  and Roth, A.
\newblock {Differential privacy: an economic method for choosing epsilon}.
\newblock In \emph{IEEE CSF}, 2014.

\bibitem[{IAPP}(2023)]{iapp2023generative}
{IAPP}.
\newblock {Generative AI: Privacy and tech perspectives}.
\newblock
  \url{https://iapp.org/news/a/generative-ai-privacy-and-tech-perspectives/},
  2023.

\bibitem[{ICO UK}(2021)]{ico2021how}
{ICO UK}.
\newblock {Chapter 2: how do we ensure anonymisation is effective?}
\newblock
  \url{https://ico.org.uk/media/about-the-ico/documents/4018606/chapter-2-anonymisation-draft.pdf},
  2021.

\bibitem[{ICO UK}(2022)]{ico2022privacy}
{ICO UK}.
\newblock {Chapter 5: privacy-enhancing technologies (PETs)}.
\newblock
  \url{https://ico.org.uk/media/about-the-ico/consultations/4021464/chapter-5-anonymisation-pets.pdf},
  2022.

\bibitem[Jordon et~al.(2018)Jordon, Yoon, and Van Der~Schaar]{jordon2018pate}
Jordon, J., Yoon, J., and Van Der~Schaar, M.
\newblock {PATE-GAN: generating synthetic data with differential privacy
  guarantees}.
\newblock In \emph{ICLR}, 2018.

\bibitem[Jordon et~al.(2022)Jordon, Szpruch, Houssiau, Bottarelli, Cherubin,
  Maple, Cohen, and Weller]{jordon2022synthetic}
Jordon, J., Szpruch, L., Houssiau, F., Bottarelli, M., Cherubin, G., Maple, C.,
  Cohen, S.~N., and Weller, A.
\newblock {Synthetic Data--what, why and how?}
\newblock \emph{arXiv:2205.03257}, 2022.

\bibitem[Kotelnikov et~al.(2022)Kotelnikov, Baranchuk, Rubachev, and
  Babenko]{kotelnikov2022tabddpm}
Kotelnikov, A., Baranchuk, D., Rubachev, I., and Babenko, A.
\newblock {TabDDPM: Modelling Tabular Data with Diffusion Models}.
\newblock \emph{arXiv:2209.15421}, 2022.

\bibitem[Kulynych et~al.(2023)Kulynych, Hsu, Troncoso, and
  Calmon]{kulynych2023arbitrary}
Kulynych, B., Hsu, H., Troncoso, C., and Calmon, F.~P.
\newblock {Arbitrary decisions are a hidden cost of differentially-private
  training}.
\newblock In \emph{ACM FAccT}, 2023.

\bibitem[L{\'o}pez \& Elbi(2022)L{\'o}pez and Elbi]{lopez2022on}
L{\'o}pez, C. A.~F. and Elbi, A.
\newblock {On the legal nature of synthetic data}.
\newblock In \emph{NeurIPS SyntheticData4ML}, 2022.

\bibitem[McSherry(2016)]{mcsherry2016statistical}
McSherry, F.
\newblock {Statistical inference considered harmful}.
\newblock
  \url{https://github.com/frankmcsherry/blog/blob/master/posts/2016-06-14.md},
  2016.

\bibitem[Nasr et~al.(2021)Nasr, Songi, Thakurta, Papernot, and
  Carlin]{nasr2021adversary}
Nasr, M., Songi, S., Thakurta, A., Papernot, N., and Carlin, N.
\newblock {Adversary instantiation: lower bounds for differentially private
  machine learning}.
\newblock In \emph{IEEE S\&P}, 2021.

\bibitem[Nissim et~al.(2017)Nissim, Bembenek, Wood, Bun, Gaboardi, Gasser,
  O'Brien, Steinke, and Vadhan]{nissim2017bridging}
Nissim, K., Bembenek, A., Wood, A., Bun, M., Gaboardi, M., Gasser, U., O'Brien,
  D.~R., Steinke, T., and Vadhan, S.
\newblock {Bridging the gap between computer science and legal approaches to
  privacy}.
\newblock \emph{Harvard JOLT}, 2017.

\bibitem[{ONS UK}(2018)]{ons2018privacy}
{ONS UK}.
\newblock {Privacy and data confidentiality methods: a data and analysis method
  review}.
\newblock
  \url{https://analysisfunction.civilservice.gov.uk/policy-store/privacy-and-data-confidentiality-methods-a-national-statisticians-quality-review-nsqr/},
  2018.

\bibitem[Papernot et~al.(2017)Papernot, Abadi, Erlingsson, Goodfellow, and
  Talwar]{papernot2017semi}
Papernot, N., Abadi, M., Erlingsson, U., Goodfellow, I., and Talwar, K.
\newblock {Semi-supervised knowledge transfer for deep learning from private
  training data}.
\newblock In \emph{ICLR}, 2017.

\bibitem[Papernot et~al.(2018)Papernot, Song, Mironov, Raghunathan, Talwar, and
  Erlingsson]{papernot2018scalable}
Papernot, N., Song, S., Mironov, I., Raghunathan, A., Talwar, K., and
  Erlingsson, {\'U}.
\newblock {Scalable private learning with pate}.
\newblock In \emph{ICLR}, 2018.

\bibitem[{Politico}(2023)]{politico2023italian}
{Politico}.
\newblock {Italian privacy regulator bans ChatGPT}.
\newblock
  \url{https://www.politico.eu/article/italian-privacy-regulator-bans-chatgpt/},
  2023.

\bibitem[{Reuters}(2023)]{reuters2023chatgpt}
{Reuters}.
\newblock {ChatGPT sets record for fastest-growing user base}.
\newblock
  \url{https://www.reuters.com/technology/chatgpt-sets-record-fastest-growing-user-base-analyst-note-2023-02-01/},
  2023.

\bibitem[{Royal Society}(2023)]{rs2023privacy}
{Royal Society}.
\newblock {From privacy to partnership: the role of PETs in data governance and
  collaborative analysis}.
\newblock
  \url{https://royalsociety.org/-/media/policy/projects/privacy-enhancing-technologies/From-Privacy-to-Partnership.pdf},
  2023.

\bibitem[{Sequoia Capital}(2022)]{sequoia2022generative}
{Sequoia Capital}.
\newblock {Generative AI: A Creative New World}.
\newblock
  \url{https://www.sequoiacap.com/article/generative-ai-a-creative-new-world/},
  2022.

\bibitem[Sohl-Dickstein et~al.(2015)Sohl-Dickstein, Weiss, Maheswaranathan, and
  Ganguli]{sohl2015deep}
Sohl-Dickstein, J., Weiss, E., Maheswaranathan, N., and Ganguli, S.
\newblock {Deep unsupervised learning using nonequilibrium thermodynamics}.
\newblock In \emph{ICML}, 2015.

\bibitem[Solatorio \& Dupriez(2023)Solatorio and
  Dupriez]{solatorio2023realtabformer}
Solatorio, A.~V. and Dupriez, O.
\newblock {REaLTabFormer: Generating Realistic Relational and Tabular Data
  using Transformers}.
\newblock \emph{arXiv:2302.02041}, 2023.

\bibitem[Stadler et~al.(2022)Stadler, Oprisanu, and
  Troncoso]{stadler2022synthetic}
Stadler, T., Oprisanu, B., and Troncoso, C.
\newblock {Synthetic data -- anonymization groundhog day}.
\newblock In \emph{Usenix Security}, 2022.

\bibitem[{TechCrunch}(2023{\natexlab{a}})]{techcrunch2023the}
{TechCrunch}.
\newblock {The current legal cases against generative AI are just the
  beginning}.
\newblock
  \url{https://techcrunch.com/2023/01/27/the-current-legal-cases-against-generative-ai-are-just-the-beginning/},
  2023{\natexlab{a}}.

\bibitem[{TechCrunch}(2023{\natexlab{b}})]{techcrunch2023vcs}
{TechCrunch}.
\newblock {VCs continue to pour dollars into generative AI}.
\newblock
  \url{https://techcrunch.com/2023/03/28/generative-ai-venture-capital/},
  2023{\natexlab{b}}.

\bibitem[{UN}(2023)]{un2023guide}
{UN}.
\newblock {The United Nations Guide on privacy-enhancing technologies for
  official statistics}.
\newblock
  \url{https://unstats.un.org/bigdata/task-teams/privacy/guide/2023_UN%20PET%20Guide.pdf},
  2023.

\bibitem[van~den Burg \& Williams(2021)van~den Burg and
  Williams]{van2021memorization}
van~den Burg, G. and Williams, C.
\newblock {On memorization in probabilistic deep generative models}.
\newblock \emph{NeurIPS}, 2021.

\bibitem[Vaswani et~al.(2017)Vaswani, Shazeer, Parmar, Uszkoreit, Jones, Gomez,
  Kaiser, and Polosukhin]{vaswani2017attention}
Vaswani, A., Shazeer, N., Parmar, N., Uszkoreit, J., Jones, L., Gomez, A.~N.,
  Kaiser, {\L}., and Polosukhin, I.
\newblock {Attention is all you need}.
\newblock \emph{NeurIPS}, 2017.

\bibitem[Xie et~al.(2018)Xie, Lin, Wang, Wang, and Zhou]{xie2018differentially}
Xie, L., Lin, K., Wang, S., Wang, F., and Zhou, J.
\newblock {Differentially private generative adversarial network}.
\newblock \emph{arXiv:1802.06739}, 2018.

\bibitem[Xu et~al.(2023)Xu, Ganev, Joubert, Davison, Van~Acker, and
  Robinson]{xu2023synthetic}
Xu, K., Ganev, G., Joubert, E., Davison, R., Van~Acker, O., and Robinson, L.
\newblock Synthetic data generation of many-to-many datasets via random graph
  generation.
\newblock In \emph{ICLR}, 2023.

\end{thebibliography}

\bibliographystyle{icml2023}
}

\end{document}